# Computational Complexity of Probabilistic Disambiguation by means of Tree-Grammars

Khalil Sima'an*
Research Institute for Language and Speech,
Utrecht University, Trans 10, 3512 JK Utrecht, The Netherlands.
Email: khalil.simaan@let.ruu.nl.

## Abstract

This paper studies the computational complexity of disambiguation under probabilistic tree-grammars as in (Bod, 1992; Schabes and Waters, 1993). It presents a proof that the following problems are NP-hard: computing the Most Probable Parse from a sentence or from a word-graph, and computing the Most Probable Sentence (MPS) from a word-graph. The NP-hardness of computing the MPS from a word-graph also holds for Stochastic Context-Free Grammars (SCFGs).

## 1 Motivation

Statistical disambiguation is currently a popular technique in parsing Natural Language. Among the models that implement statistical disambiguation one finds the models that employ Tree-Grammars such as Data Oriented Parsing (DOP) (Scha, 1990; Bod, 1992) and Stochastic (Lexicalized) Tree-Adjoining Grammar (STAG) (Schabes and Waters, 1993). These models extend the domain of locality for expressing constraints from simple Context-Free Grammar (CFG) productions to deeper structures called elementary-trees. Due to this extension, the one to one mapping between a *derivation* and a *parse-tree*, which holds in CFGs, does not hold any more; many derivations might generate the same parse-tree. This seemingly spurious ambiguity turns out crucial for statistical disambiguation as defined in (Bod, 1992) and in (Schabes and Waters, 1993), where the derivations are considered different stochastic processes and their probabilities all contribute to the probability of the generated parse. Therefore the Most Probable Derivation (**MPD**) does not necessarily generate the Most Probable Parse (**MPP**).

The problem of computing the MPP in the DOP framework was put forward in (Bod, 1995). The solution which Bod proposes is Monte-Carlo estimation (Bod, 1993), which is essentially repeated random-sampling for minimizing error-rate. A Viterbi-style optimization for computing the MPP under DOP is presented in (Sima'an et al., 1994), but it does not guarantee deterministic polynomial-time complexity. In this paper we present a proof that computing the MPP under the above mentioned stochastic tree grammars is NP-hard. Note that for computing the MPD there are deterministic polynomial-time algorithms (Schabes and Waters, 1993; Sima'an, 1996)[1]. Another problem that turns out also NP-hard is computing the Most Probable Sentence (**MPS**) from a given word-graph. But this problem turns out NP-hard even for SCFGs.

Beside the mathematical interest, this work is driven by the desire to develop efficient algorithms for these problems. Such algorithms can be useful for various applications that demand robust and faithful disambiguation e.g. Speech Recognition, Information Retrieval. This proof provides an explanation for the source of complexity, and forms a license to redirect the research for solutions towards non-standard optimizations.

The structure of the paper is as follows. Section 2 briefly discusses the preliminaries. Section 3 presents the proofs. Section 4 discusses this result, points to the source of complexity and suggests some possible solutions. The presentation is formal only where it seemed necessary.

## 2 Preliminaries

### 2.1 Stochastic Tree-Substitution Grammar (STSG)

STSGs and SCFGs are closely related. STSGs and SCFGs are equal in weak generative ca-pacity (i.e. string languages). This is not the case for strong generative capacity (i.e. tree languages); STSGs can generate tree-languages that are not generatable by SCFGs. **An STSG** is a five-tuple $(V_N, V_T, S, C, PT)$, where $V_N$ and $V_T$ denote respectively the finite set of non-terminal and terminal symbols, $S$ denotes the start non-terminal, $C$ is a finite set of elementary-trees (of arbitrary depth $\geq 1$) and $PT$ is a function which assigns a value $0 \leq PT(t) \leq 1$ (probability) to each elementary-tree $t$ such that for all $N \in V_N$: $\sum_{t \in C, root(t)=N} PT(t) = 1$ (where $root(t)$ denotes the root of tree $t$). An **elementary-tree** in $C$ has only non-terminals as internal nodes but may have both terminals and non-terminals on its frontier. A non-terminal on the frontier is called an **Open-Tree** (OT). If the left-most open-tree $N$ of tree $t$ is equal to the root of tree $t1$ then $t \circ t1$ denotes the tree obtained by substituting $t1$ for $N$ in $t$. The partial function $\circ$ is called **left-most substitution**. A **left-most derivation (l.m.d.)** is a sequence of left-most substitutions $lmd = (\ldots(t_1 \circ t_2) \circ \ldots) \circ t_n$, where $t_1, \ldots, t_n \in C$, $root(t_1) = S$ and the frontier of $lmd$ consists of only terminals. The probability $P(lmd)$ is defined as $PT(t_1) \times \ldots \times PT(t_n)$. For convenience, *derivation* in the sequel refers to l.m. derivation. A **Parse** is a tree generated by a derivation. *A parse is possibly generatable by many derivations.* The probability of a parse is defined as the sum of the probabilities of the derivations that generate it. The probability of a sentence is the sum of the probabilities of all derivations that generate that sentence.

---

*Special thanks to Christer Samuelsson who pointed out and helped in solving a problem with a previous version. Thanks to Remko Scha, Rens Bod and Eric Aarts for valuable comments, and to Steven Krauwer and the STT for the support.

[1] The author notes that the *actual* accuracy figures of the experiments listed in (Sima'an, 1995) are *much higher* than the accuracy figures reported in the paper. The lower figures reported in that paper are due to a test-procedure.

A word-graph over the alphabet $Q$ is $Q_1 \times \cdots \times Q_m$, where $Q_i \subseteq Q$, for all $1 \leq i \leq m$. We denote this word-graph with $Q^m$ if $Q_i = Q$, for all $1 \leq i \leq m$.

### 2.2 The 3SAT problem

It is sufficient to prove that a problem is NP-hard in order to prove that it is intractable. A problem is NP-hard if it is (at least) as hard as any problem that has been proved to be NP-complete (i.e. a problem that is known to be decidable on a non-deterministic Turing Machine in polynomial-time but not known to be decidable on a deterministic Turing Machine in polynomial-time). To prove that problem A is as hard as problem B, one shows a reduction from problem B to problem A. The reduction must be a deterministic polynomial time transformation that preserves answers.

The NP-complete problem which forms our starting-point is the 3SAT (satisfiability) problem. An instance INS of 3SAT can be stated as follows[2]:

> Given an arbitrary[3] Boolean formula in 3-conjunctive normal form (3CNF) over the variables $u_1, \ldots, u_n$. Is there an assignment of values **true** or **false** to the Boolean variables such that the given formula is true ? Let us denote the given formula by $C_1 \wedge C_2 \wedge \cdots \wedge C_m$ for $m \geq 1$ where $C_i$ represents $(d_{i1} \vee d_{i2} \vee d_{i3})$, for $1 \leq i \leq m$, $1 \leq j \leq 3$, and $d_{ij}$ represents a literal $u_k$ or $\overline{u}_k$ for some $1 \leq k \leq n$.

Optimization problems are known to be (at least) as hard as their decision counterparts (Garey and Johnson, 1981). The decision problem related to maximizing a quantity M which is a function of a variable V can be stated as follows: is there a value for V that makes the quantity M greater than or equal to a predetermined value m. The decision problems related to disambiguation under DOP can be stated as follows, where G is an STSG, $WG$ is a word-graph, $w_0^n$ is a sentence and $0 < p \leq 1$:

**MPPWG** Does the word-graph $WG$ have any parse, generatable by the STSG G, that has probability value greater than or equal to $p$ ?

**MPS** Does the word-graph $WG$ contain any sentence, generatable by the STSG G, that has probability value greater than or equal to $p$ ?

**MPP** Does the sentence $w_0^n$ have a parse generatable by the STSG G, that has probability value greater than or equal to $p$ ?

Note that in the sequel MPPWG / MPS / MPP denotes the *decision* problem corresponding to the problem of computing the MPP / MPS / MPP from a word-graph / word-graph / sentence respectively.

## 3 Complexity of MPPWG, MPS and MPP

### 3.1 3SAT to MPPWG and MPS

The reduction from the 3SAT instance INS to an MPPWG problem must construct an STSG and a word-graph in deterministic polynomial-time. Moreover, the answers to the MPPWG instance must correspond exactly to the answers to INS. The presentation of the reduction shall be accompanied by an example of the following 3SAT instance (Barton et al., 1987): $(u_1 \vee \overline{u}_2 \vee u_3) \wedge (\overline{u}_1 \vee u_2 \vee \overline{u}_3)$. Note that a 3SAT instance is satisfiable iff at least one of the literals in each conjunct is assigned the value True. Implicit in this, but crucial, the different occurrences of the literals of the same variable must be assigned values *consistently*.

**Reduction:** The reduction constructs an STSG and a word-graph. The STSG has start-symbol $S$, two terminals represented by **T** and **F**, non-terminals which include (beside $S$) all $C_k$, for

---

[2] In the sequel, INS, INS's formula and its symbols refer to this particular instance of 3SAT.

[3] Without loss of generality we assume that the formula does not contain repetition of conjuncts.

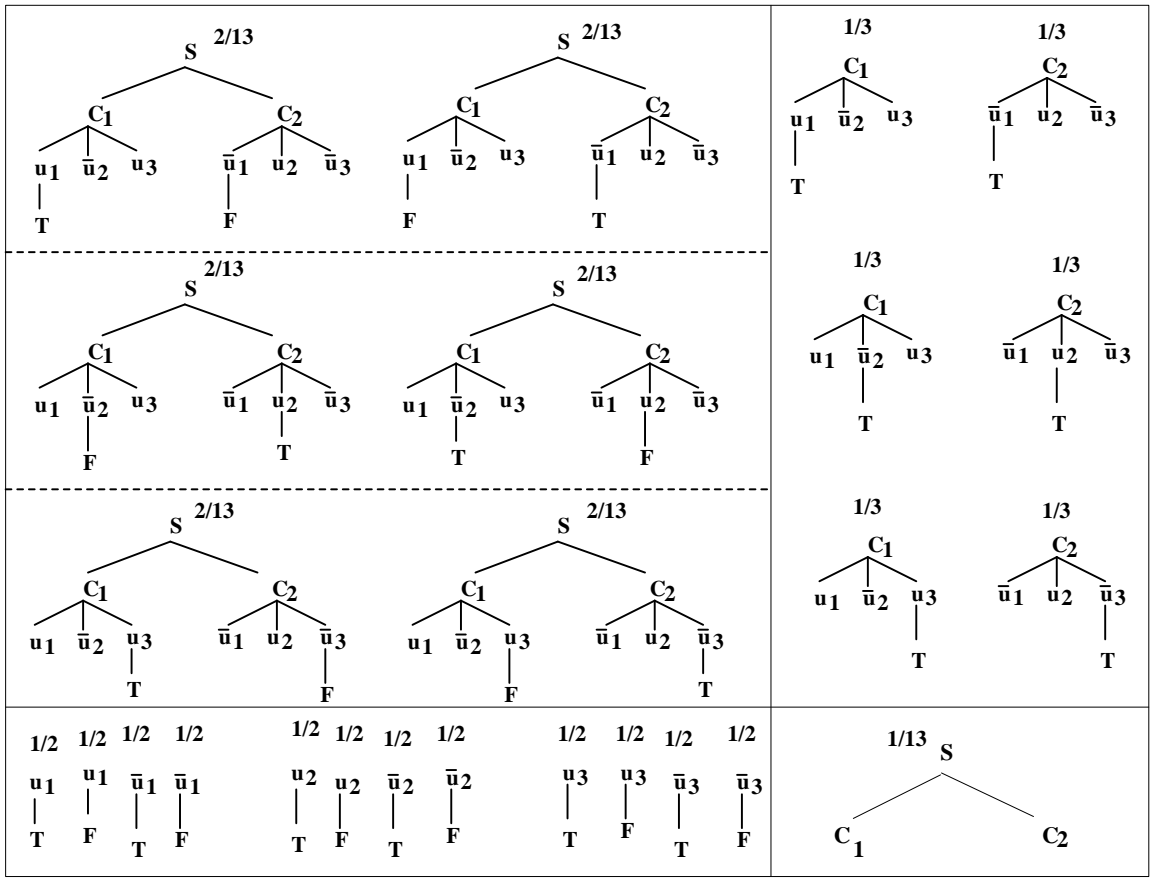

Figure 1: The elementary-trees for the example 3SAT instance

$1 \leq k \leq m$, and both literals of each Boolean variable of the formula of INS. The set of elementary-trees and probability function and the word-graph are constructed as follows:

1. For each Boolean variable $u_i$, $1 \leq i \leq n$, construct two elementary-trees that correspond to assigning the values true and false to $u_i$ *consistently* through the whole formula. Each of these elementary-trees has root $S$, with children $C_k$, $1 \leq k \leq m$, in the same order as these appear in the formula of INS; subsequently the children of $C_k$ are the non-terminals that correspond to its three disjuncts $d_{k1}$, $d_{k2}$ and $d_{k3}$. And finally, the assignment of true (false) to $u_i$ is modeled by creating a child terminal **T** (resp. **F** ) to each non-terminal $u_i$ and **F** (resp. **T** ) to each $\overline{u}_i$. The two elementary-trees for $u_1$, of our example, are shown in the top left corner of figure 1.

2. The reduction constructs three elementary-trees for each conjunct $C_k$. The three elementary-trees for conjunct $C_k$ have the same internal structure: root $C_k$, with three children that correspond to the disjuncts $d_{k1}$, $d_{k2}$ and $d_{k3}$. In each of these elementary-trees exactly one of the disjuncts has as a child the terminal **T** ; in each of them this is a different one. Each of these elementary-trees corresponds to the conjunct where one of the three possible literals is assigned the value **T** . For the elementary-trees of our example see the top right corner of figure 1.

3. The reduction constructs for each of the two literals of each variable $u_i$ two elementary-trees where the literal is assigned in one case **T** and in the other **F** . Figure 1 shows these elementary-trees for variable $u_1$ in the bottom left corner.

4. The reduction constructs one elementary-tree that has root $S$ with children $C_k$, $1 \leq k \leq m$, in the same order as these appear in the formula of INS (see the bottom right corner of figure 1).

5. The probabilities of the elementary-trees that have the same root non-terminal sum up to 1. The probability of an elementary-tree with root $S$ that was constructed in step 1 of this reduction is a value $p_i$, $1 \leq i \leq n$, where $u_i$ is the only variable of which the literals in the elementary-tree at hand are lexical-ized (i.e. have terminal children). Let $n_i$ denote the number of occurrences of both literals of variable $u_i$ in the formula of INS. Then $p_i = \theta \left(\frac{1}{2}\right)^{n_i}$, for some real $\theta$ that has to fulfill some conditions which will be derived next. The probability of the tree rooted with $S$ and constructed at step 4 of this reduction must then be $p_0 = [1 - 2\sum_{i=1}^{n} p_i]$. The probability of the elementary-trees of root $C_k$ (step 2) is $\left(\frac{1}{3}\right)$, and of root $u_i$ or $\overline{u}_i$ (step 3) is $\left(\frac{1}{2}\right)$. For our example some suitable probabilities are shown in figure 1.

6. Let $Q$ denote a threshold probability that shall be derived hereunder. The MPPWG (MPS) instance is: *does the STSG generate a parse (resp. sentence) of probability $\geq Q$, for the word-graph $WG = \{\mathbf{T}, \mathbf{F}\}^{3\mathbf{m}}$ ?*

**Deriving the probabilities:** The parses generated by the constructed STSG differ only in the sentences on their frontiers. Therefore, if a sentence is generated by this STSG then it has exactly one parse. This justifies the choice to reduce 3SAT to MPPWG and MPS simultaneously.

One can recognize two types of derivations in this STSG. The **first** type corresponds to substituting for an open-tree (i.e literal) of any of the $2n$ elementary-trees constructed in step 1 of the reduction. This type of derivation corresponds to assigning values to all literals of some variable $u_i$ in a consistent manner. For all $1 \leq i \leq n$ the probability of a derivation of this type is

$$p_i \left(\frac{1}{2}\right)^{3m-n_i} = \theta \left(\frac{1}{2}\right)^{3m}$$

The **second** type of derivation corresponds to substituting the elementary-trees rooted with $C_k$ in $S \rightarrow C_1 \ldots C_m$, and subsequently substituting in the open-trees that correspond to literals. This type of derivation corresponds to assigning to at least one literal in each conjunct the value true. The probability of any such derivation is

$$p_0 \left(\frac{1}{2}\right)^{2m} \left(\frac{1}{3}\right)^m = [1 - 2\theta \sum_{i=1}^{n} \left(\frac{1}{2}\right)^{n_i}] \left(\frac{1}{2}\right)^{2m} \left(\frac{1}{3}\right)^m$$

Now we derive both the threshold $Q$ and the parameter $\theta$. Any parse (or sentence) that fulfills both the "consistency of assignment" requirements and the requirement that each conjunct has at least one literal with child $\mathbf{T}$, must be generated by $n$ derivations of the first type and at least one derivation of the second type. Note that a parse can never be generated by more than $n$ derivations of the first type. Thus the threshold $Q$ is:

$$Q = n\theta \left(\frac{1}{2}\right)^{3m} + [1 - 2\theta \sum_{i=1}^{n} \left(\frac{1}{2}\right)^{n_i}] \left(\frac{1}{2}\right)^{2m} \left(\frac{1}{3}\right)^m$$

However, $\theta$ must fulfill some requirements for our reduction to be acceptable:

1. For all i: $0 < p_i < 1$. This means that for $1 \leq i \leq n$: $0 < \theta \left(\frac{1}{2}\right)^{n_i} < 1$, and $0 < p_0 < 1$. However, the last requirement on $p_0$ implies that $0 < 2\theta \sum_{i=1}^{n} \left(\frac{1}{2}\right)^{n_i} < 1$, which is a stronger requirement than the other $n$ requirements. This requirement can also be stated as follows: $0 < \theta < \frac{1}{2\sum_{i=1}^{n}(\frac{1}{2})^{n_i}}$.

2. Since we want to be able to know whether a parse is generated by a second type derivation only by looking at the probability of the parse, the probability of a second type derivation must be distinguishable from first type derivations. Moreover, if a parse is generated by more than one derivation of the second type, we do not want the sum of the probabilities of these derivations to be mistaken for one (or more) first type derivation(s). For any parse, there are at most $3^m$ second type derivations (e.g. the sentence $\mathbf{T} \ldots \mathbf{T}$). Therefore we require that:

$$3^m [1 - 2\theta \sum_{i=1}^{n} \left(\frac{1}{2}\right)^{n_i}] \left(\frac{1}{2}\right)^{2m} \left(\frac{1}{3}\right)^m < \theta \left(\frac{1}{2}\right)^{3m}$$

Which is equal to $\theta > \frac{1}{2\sum_{i=1}^{n}(\frac{1}{2})^{n_i} + (\frac{1}{2})^m}$.

3. For the resulting STSG to be a probabilistic model, the "probabilities" of parses and sentences must be in the interval $(0, 1]$. This is taken care of by demanding that the sum of the probabilities of elementary-trees that have the same root non-terminal is 1, and by the definition of the derivation's probability, the parse's probability, and the sentence's probability.

There exists a $\theta$ that fulfills all these requirements because the lower bound $\frac{1}{2\sum_{i=1}^{n}(\frac{1}{2})^{n_i} + (\frac{1}{2})^m}$ is always larger than zero and is strictly smaller than the upper bound $\frac{1}{2\sum_{i=1}^{n}(\frac{1}{2})^{n_i}}$.

**Polynomiality of the reduction:** This reduction is deterministic polynomial-time in $n$ because it constructs not more than $2n + 1 + 3m + 4n$ elementary-trees of maximum number of nodes[4] $7m + 1$.

**The reduction preserves answers:** The proof concerns the only two possible answers.

Yes If INS's answer is Yes then there is an assignment to the variables that is consistent and where each conjunct has at least one literal assigned true. Any possible assignment is represented by one sentence in $WG$. A sentence which corresponds to a "successful" assignment must be generated by $n$ derivations of the first type and at least one derivation of the second type; this is because the sentence $w_1^{3m}$ fulfills $n$ consistency requirements (one per Boolean variable) and has at least one **T** as $w_{3k+1}$, $w_{3k+2}$ or $w_{3k+3}$, for all $0 \leq k < m$. Both this sentence and its corresponding parse have probability $\geq Q$. Thus MPPWG and MPS also answer Yes.

---

[4] Note than $m$ is polynomial in $n$ because the formula does not contain two identical conjuncts.

No If INS's answer is No, then all possible assignments are either not consistent or result in at least one conjunct with three false disjuncts, or both. The sentences (parses) that correspond to non-consistent assignments each have a probability that cannot result in a Yes answer. This is the case because such sentences have fewer than $n$ derivations of the first type, and the derivations of the second type can never compensate for that (the requirements on $\theta$ take care of this). For the sentences (parses) that correspond to consistent assignments, there is at least some $0 \leq k < m$ such that $w_{3k+1}$, $w_{3k+2}$ and $w_{3k+3}$ are all **F**. These sentences do not have second type derivations. Thus, there is no sentence (parse) that has a probability that can result in a Yes answer; the answer of MPPWG and MPS is NO.

We conclude that MPPWG and MPS are both NP-hard problems.

Now we show that MPPWG and MPS are in NP. A problem is in NP if it is decidable by a non-deterministic Turing machine. The proof here is informal: we show a non-deterministic algorithm that keeps proposing solutions and then checking each of them in deterministic polynomial time cf. (Barton et al., 1987). If one solution is successful then the answer is Yes. One possible non-deterministic algorithm for the MPPWG and MPS, constructs firstly a parse-forest for $WG$ in deterministic polynomial time based on the algorithms in (Schabes and Waters, 1993; Sima'an, 1996), and subsequently traverses this parse-forest (bottom-up for example) deciding at each point what path to take. Upon reaching the start non-terminal $S$, it retrieves the sentence (parse) and evaluates it in deterministic polynomial-time (Sima'an et al., 1994), thereby answering the decision problem.

This concludes the proof that MPPWG and MPS are both NP-complete.

### 3.2 NP-completeness of MPP

The NP-completeness of MPP can be easily deduced from the previous section. In the reduction the terminals of the constructed STSG are new symbols $v_{ij}$, $1 \leq i \leq m$ and $1 \leq j \leq 3$, instead of **T** and **F** that become non-terminals. Each of the elementary-trees with root $S$ or $C_k$ is also represented here but each **T** and **F** on the frontier has a child $v_{kj}$ wherever the **T** or **F** appears as the child of the $j$th child (a literal) of $C_k$. For each elementary-tree with root $u_i$ or $\overline{u}_i$, there are $3m$ elementary-trees in the new STSG that correspond each to creating a child $v_{ij}$ for the **T** or **F** on its frontier. The probability of an elementary-tree rooted by a literal is $\frac{1}{6m}$. The probabilities of elementary-trees rooted with $C_k$ do not change. And the probabilities of the elementary-trees rooted with $S$ are adapted from the previous reduction by substituting for every $(\frac{1}{2})$ the value $\frac{1}{6m}$. The threshold $Q$ and the requirements on $\theta$ are also updated accordingly. The input sentence which the reduction constructs is simply $v_{11} \ldots v_{3m}$. The decision problem is *whether there is a parse generated by the resulting STSG for this sentence that has probability larger than or equal to $Q$.*

The rest of the proof is very similar to that in section 3. Therefore the decision problem MPP is NP-complete.

### 3.3 MPS under SCFG

The decision problem MPS is NP-complete also under SCFG. The proof is easily deducible from the proof concerning MPS for STSGs. The reduction simply takes the elementary-trees of the MPS for STSGs and removes their internal structure, thereby obtaining simple CFG productions. Crucially, each elementary-tree results in one unique CFG production. The probabilities are kept the same. The word-graph is also the same word-graph as in MPS for STSGs. The problem is: *does the SCFG generate a sentence with probability $\geq Q$, for the word-graph $WG = \{\mathbf{T}, \mathbf{F}\}^{3m}$.* The rest of the proof follows directly from section 3.

## 4 Conclusion and discussion

We conclude that computing the MPP / MPS / MPP from a sentence / word-graph / word-graph respectively is NP-hard under DOP. Computing the MPS from a word-graph is NP-hard even under SCFGs. Moreover, these results are applicable to STAG as in (Schabes and Waters, 1993).

The proof of the previous section helps in understanding why computing the MPP in DOP is such a hard problem. The fact that MPS under SCFG is also NP-hard implies that the complexity of the MPPWG, MPS and MPP is due to the definitions of the probabilistic model rather than the complexity of the syntactic model.

The main source of NP-completeness is the following common structure of these problems: they all search for an entity that maximizes the *sum* of the probabilities of processes which depend on that entity. For the MPS problem of SCFGs for example, one searches for the sentence which maximizes the sum of the probabilities of *the parses that generate that sentence* (i.e. the probability of a parse is also a function of whether it generates the sentence at hand or not). This is not the case, for example, when computing the MPD under STSGs (for sentence or even a word-graph), or when computing the MPP under SCFGs (for a sentence or a word-graph).

The proof in this paper is not a mere theoretical issue. An exponential algorithm can be comparable to a deterministic polynomial algorithm if the grammar-size can be neglected and if the exponential formula is not much worse than the polynomial for realistic sentence lengths. But as soon as the grammar size becomes an important factor (e.g. in DOP), polynomiality becomes a very desirable quality. For example $|G| e^n$ and $|G| n^3$ for $n \leq 7$ are comparable but for $n = 12$ the polynomial is some 94 times faster. If the grammar size is small and the comparison is between 0.001 seconds and 0.1 seconds this might be of no practical importance. But when the grammar size is large and the comparison is between 60 seconds[5] and 5640 seconds for a sentence of length 12, then things become different.

To compute the MPP under DOP, one possible solution involves some heuristic that directs the search towards the MPP; a form of this strategy is the Monte-Carlo technique. Another solution might involve assuming Memory-based behavior in directing the search towards the most "suitable" parse according to some heuristic evaluation function that is inferred from the probabilistic model. And a third possible solution is to adjust the probabilities of elementary-trees such that it is not necessary to compute the MPP. The probability of an *elementary-tree* can be redefined as the sum of the probabilities of all derivations that generate it in the given STSG. This redefinition can be applied by off-line computation and normalization. Then the probability of a parse is redefined as the probability of the MPD that generates it, thereby collapsing the MPP and MPD. This method assumes full independence beyond the borders of elementary-trees, which might be an acceptable assumption.

Finally, it is worth noting that the solutions that we suggested above are merely algorithmic. But the ultimate solution to the complexity of probabilistic disambiguation under the current models lies, we believe, only in further incorporation of the crucial elements of the human processing ability into these models.

---

[5] This is a realistic figure from experiments on the ATIS.